\documentclass{article} % For LaTeX2e
\usepackage{iclr2017_conference,times}
\usepackage{hyperref}
\usepackage{url}
\usepackage{bbding}
\usepackage{pifont}
\usepackage{wasysym}
\usepackage{amssymb}
\usepackage{graphicx} % removed [dvips] as argument to fix figure issue, check back if further issues arise
\usepackage{siunitx} % removed [output-decimal-marker={,},exponent-product=\cdot]

\title{Adaptive Blending Units: Trainable Activation Functions for Deep Neural Networks}

\author{Leon Ren\'e S\"utfeld, Flemming Brieger, Holger Finger, Sonja F\"ullhase \& Gordon Pipa \\
Institute for Cognitive Science\\
Osnabr\"uck University\\
Wachsbleiche 27, 49090 Osnabr\"uck, Germany \\
\texttt{\{lsuetfel,fbrieger,hofinger,sfuellhase,gpipa\}@uos.de} \\
}

\begin{document}

\maketitle

\begin{abstract}
The most widely used activation functions in current deep feed-forward neural networks are rectified linear units (ReLU), and many alternatives have been successfully applied, as well. However, none of the alternatives have managed to consistently outperform the rest and there is no unified theory connecting properties of the task and network with properties of activation functions for most efficient training. A possible solution is to have the network learn its preferred activation functions. In this work, we introduce Adaptive Blending Units (ABUs), a trainable linear combination of a set of activation functions. Since ABUs learn the shape, as well as the overall scaling of the activation function, we also analyze the effects of adaptive scaling in common activation functions. We experimentally demonstrate advantages of both adaptive scaling and ABUs over common activation functions across a set of systematically varied network specifications. We further show that adaptive scaling works by mitigating covariate shifts during training, and that the observed advantages in performance of ABUs likewise rely largely on the activation function's ability to adapt over the course of training.
\end{abstract}

\section{Introduction}

Deep neural networks are structured around layers, each of which performs a linear transformation of its input before feeding the signal through a scalar non-linear activation function. Chaining larger numbers of non-linear functions then allows the networks to find and extract complex features in the input. Activation functions thus have a key function in deep neural networks: Without intermittent non-linearities, these networks could only perform linear operations on the input. But despite a large number of activation functions proven successful in the literature, it remains unclear, what properties of an activation function are most desirable, given a particular task and network configuration. Ideally, the network would sort this issue out by itself, but most common activation functions are fixed during training, i.e., their shape and scaling are treated as hyperparameters. We suggest changing this practice by making an activation function's shape and scaling a trainable parameter of the network. Our main contribution in this work is the Adaptive Blending Unit (ABU), a linear combination of a set of basic activation functions that allows the shape and scaling of the resulting activation function to be learned during training. In an effort to separate and understand the effects of the activation function's shape and its scaling, we also examine the effect of adaptive scaling on common activation functions without adaptation of the shape, as well as normalizing the blending weights in ABUs, thus learning its shape without learning any scaling. Throughout this work, we apply one scaling weight, or one set of blending weights (i.e., one ABU) per layer of the network. This way, the network is free to learn the activation function and / or scaling that best suits the computations performed in any given layer, while the number of parameters in the network is kept low enough, as not to require regularization. The remainder of this work is structured as follows. In section $2$, we will review related work, before comparing and analyzing common activation functions, their adaptively scaled counterparts and ABUs on CIFAR image classification tasks in section $3$. In section $4$, we examine multiple ways of normalizing ABUs, to provide an account of adaptive shape without adaptive scaling. Finally, in section $5$, we examine pre-training of the scaling and blending weights, to examine the role of adaptiveness over the course of training.

\section{Related Work}
The most prevalent activation function in modern neural networks is the \emph{Rectified Linear Unit (ReLU)} \citep{hahnloser2000digital, nair2010rectified}, a piecewise-linear function returning the identity for all positive inputs and zero otherwise. Its constant derivative of $1$ on the positive part helps alleviating the vanishing gradient problem \citep{glorot2011deep}, making it the first activation function allowing for a large number of stacked layers to be trained efficiently. With this, ReLU was partly responsible for the breakthrough of deep neural networks around 2012, marked by AlexNet's victory in the annual ILSVRC challenge \citep{krizhevsky2012imagenet}. \emph{Leaky ReLU (LReLU)} \citep{maas2013rectifier}, \emph{Parametric ReLU (PReLU)} \citep{he2015delving}, and \emph{Randomized Leaky ReLU (RReLU)} \citep{xu2015empirical} are all based on ReLU, but replace the zero-output for negative values by a linear function. In PReLU, the slope of the negative part of the function is controlled by a trainable parameter. Exponential Linear Units (ELU) \citep{clevert2015fast} like ReLU, return the identity for positive values, but $\alpha(\exp(x)-1)$ for negative values, with $\alpha$ typically set to $1$. Scaled Exponential Linear Units (SELU) \citep{klambauer2017self} are identical to ELU, except for an additional scaling parameter $\lambda$ acting upon the function as a whole. The values for $\alpha$ and $\lambda$ in SELUs are analytically derived to ensure convergence of activations towards unit mean and variance across layers. In a more empirical approach, \cite{ramachandran2017searching} performed a large reinforcement learning-based search for successful activation functions, and found multiple novel and well-performing functions. The most successful, given by $f(x)=x\cdot\textmd{sigmoid}(\beta x)$ and named \emph{Swish}, uses the trainable parameter $\beta$ to control the overall shape of the function. E-Swish \citep{alcaide2018eswish} ditches this parameter (setting it to $1$), and instead scales the whole function by a manually determined parameter between $1$ and $2$. In addition to these, there are numerous approaches in which the activation function's overall shape is learned, often using multiple parameters: \emph{Adaptive Piecewise Linear Units (APL)} learn the slope of all piecewise linear elements and the position of the hinges independently for each neuron via backpropagation, while the number of linear pieces is a hyperparameter that is set manually \citep{agostinelli2014learning}. Similarly, Maxout activations \citep{goodfellow2013maxout} learn a convex piecewise-linear function by returning the maximum of a fixed set of neurons, while the regular network weights determine the shape of the resulting function. \cite{eisenach2016nonparametrically} use Fourier series basis expansion to approximate non-linear parameterized basis functions, and train one activation function per feature map in a convolutional network. An approach suggested by \cite{godfrey2015continuum}, called the \emph{soft exponential function}, can switch between a large number of different mathematical operations, such as addition, multiplication and exponentiation, by adjusting a trainable parameter. However, to our knowledge, no empirical validation of the approach was offered so far. In an approach similar to ours, \cite{dushkoff2016adaptive} suggested blending a set of activation functions on a per-neuron basis, and constraining the blending weights to values between $0$ and $1$, by gating them with exponential sigmoid functions. Blending activation functions on a per-neuron basis, however, required downscaling of gradient updates as a form of regularization. Most similar to our approach, \citep{manessi2018learning} suggested a learned blending of multiple common activation functions per layer, where the blending weights are constrained to sum up to $1$, and showed this approach to be successful over a range of tasks and network configurations. We will provide further details with respect to the similarities between their approach and ours in the appropriate sections.

\section{Adaptive scaling and Adaptive Blending Units}

% They each consist of $\num{60000}$ RGB images measuring $32\times 32$ pixels in size. In each case, $\num{10000}$ images are reserved for the test set, while the remaining $\num{50000}$ images are used for training. CIFAR 10 contains 10 classes of images, i.e., $\num{5000}$ images per class in the training set, whereas CIFAR 100 contains 100 classes, i.e., 500 images per class in the training set.
% Note that all design choices made for this network were made within the scope of a four-layer convolutional network and based on performance tests using only ReLU activation functions. No optimizations favoring scaled activation functions or ABUs was performed.
% He init, i.e., from a truncated normal distribution with $\mu=0$ and $\sigma=\sqrt{\frac{2}{N}}$. For convolutional filters, $N=k^2\cdot c$, where $k$ is the spatial filter size and $c$ the number of channels in the previous layer. For dense layers, $N$ is the number of weights in the weight matrix.
% Since we used mini-batches also in validation and testing, no moving average had to be used in the batch normalization.

In this section, we will introduce ABUs as an extension to the idea of an adaptive scaling of common activation functions, and analyze both ABUs and adaptive scaling with respect to task performance and the mechanics they introduce to the network. The activation functions we used as a baseline throughout this work are the \emph{hyperbolic tangent (tanh)}, \emph{ReLU}, \emph{ELU}, \emph{SELU}, the \emph{identity} and \emph{Swish}. We will reference the adaptively scaled versions of these by adding "$\alpha$" to the function's name, e.g., "$\alpha$ReLU".

\subsection{methods}

\begin{figure}[t]
\centering
\includegraphics[width=1.0\textwidth]{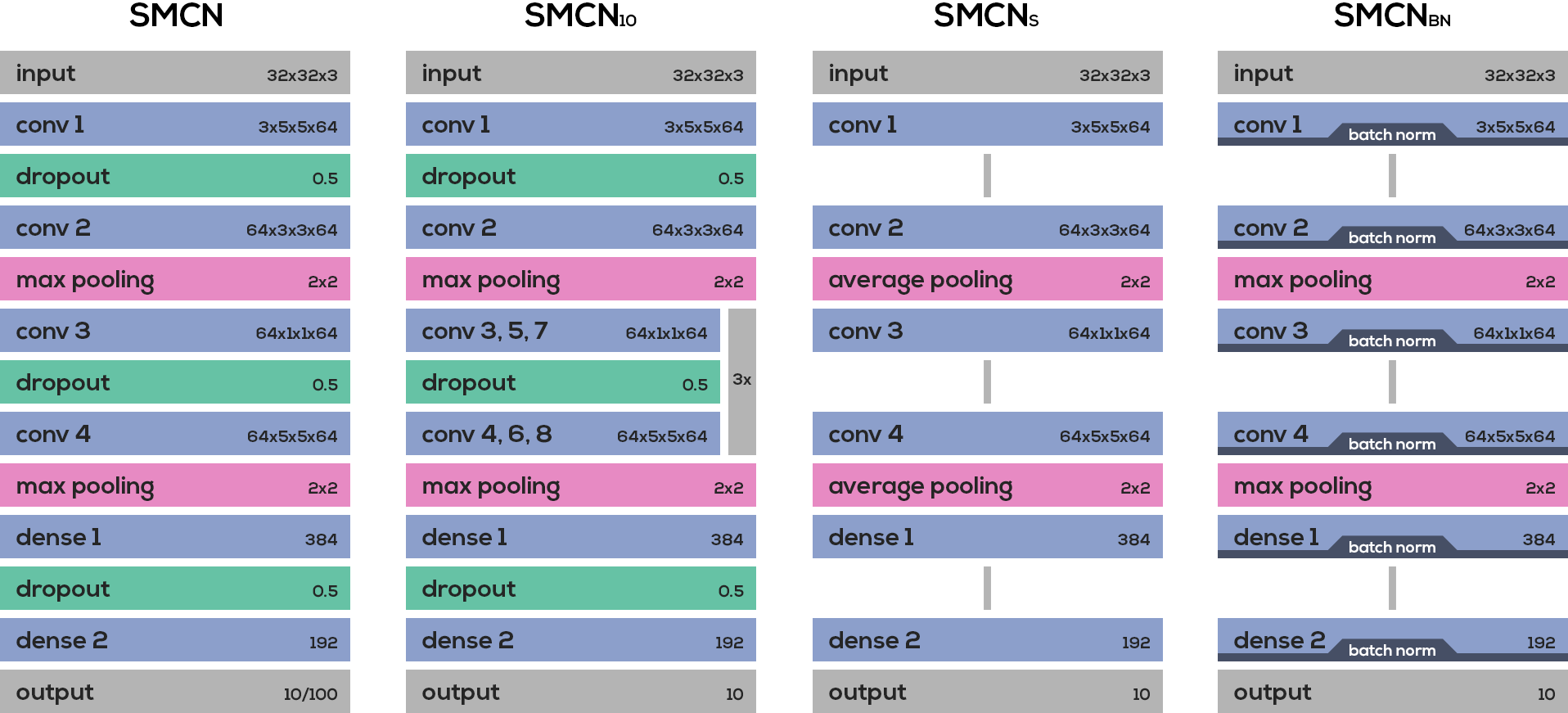}
%\fbox{\rule[-.5cm]{0cm}{4cm} \rule[-.5cm]{4cm}{0cm}}
\caption{SMCN network architectures for CIFAR10 / CIFAR100. \textbf{(Vanilla) SMCN}: Default architecture, 6 hidden layers. \textbf{SMCN\textsubscript{10}}: Medium-sized network, 10 hidden layers. \textbf{SMCN\textsubscript{S}}: Additional architecture for robustness tests; 6 hidden layers, omitting dropout layers and replacing max pooling with average pooling. \textbf{SMCN\textsubscript{BN}}: Additional architecture for robustness tests; 6 hidden layers, batch normalization after each activation, and omitting dropout layers.}
\end{figure}

Given a deep neural network of $n$ layers, and an activation function $f(x)$, the adaptively scaled version of the activation function is given by $\alpha_i\cdot f(x)$, with $i=1,\dots,n$. The \emph{scaling weights} $\alpha_i$ are initialized at $1$ by default, and trained via backpropagation alongside all other network parameters. Swish's $\beta$ is initialized as a trainable parameter per layer (i.e., $\beta_i$) and likewise trained via backpropagation in all cases. ABUs can be viewed as an extension to this approach, in which the shape of the activation function is determined by a blending of multiple common activation functions within the unit. Given a deep neural network of $n$ layers, and a set of $m$ activation functions per layer, the ABU for the $i$th layer is defined as $g_i(x) = \sum_j\alpha_{ij}\cdot f_j(x)$ with $i=1,\dots,n$ and $j=1,\dots,m$. The \emph{blending weights} $\alpha_{ij}$ are initialized at $\frac{1}{m}$ by default, and also trained via backpropagation alongside all other network parameters. With respect to the set of activation functions used in ABUs, we chose tanh, ELU, ReLU, the identity, and Swish in order to allow for high flexibility of the resulting function. However, we did not conduct an exhaustive search over possible sets of activation functions, so other sets may outperform the chosen configuration.

The \emph{CIFAR 10} and \emph{CIFAR 100} datasets \citep{krizhevsky2009learning} served as benchmarks to assess the performance of our approaches. Per-image z-transformation was applied as pre-processing to all images, and $5\%$ of the  training set was used as a validation set during training. To evaluate model performance, we applied post-hoc early stopping: The model was saved once every $8$ epochs and the validation accuracy was estimated frequently over the course of training. All networks were trained for $\num{60000}$ steps, after which we smoothed the validation accuracy curve and selected the model save point for which said curve indicated the highest performance. For each network and task specification, we report the mean of 30 runs, as well as the standard error. Training, validation and testing were all performed using mini-batches.

For the networks used in our tests, we created a set of small to mid-sized convolutional networks, called Simple Modular Convolutional Networks (SMCN). In different variations of these, features were added or subtracted to test the robustness of our approaches across different network design choices. The vanilla SMCN consists of four convolutional layers, followed by two dense layers. Max pooling ($3\times 3$, stride $2$) is performed after the second and fourth convolutional layer, and dropout \citep{srivastava2014dropout} with a rate of $0.5$ is applied after the first and third convolutional layer, and after the first dense layer. The convolutional layers use zero-padding and stride $1$. They feature filters of size $[5\times 5\times 64]$, $[3\times 3\times 64]$, $[1\times 1\times 64]$, and $[5\times 5\times 64]$, and the dense layers consist of 384 and 192 neurons, respectively (see Figure 1). The network contains no residual connections, and no batch normalization \citep{ioffe2015batch} is performed by default. Initial weights are randomly sampled using He initialization \citep{he2015delving}. Bias units were initialized at $0.1$, except for the first convolutional layer ($0.0$). The network is trained for $\num{60000}$ steps on mini-batches of size $256$, using the Adam optimizer \citep{kingma2014adam} with a learning rate of $\eta=0.001$. The total number of trainable parameters in the vanilla SMCN is roughly $1.8$M. In addition to this, we used the following variations in our tests: \emph{SMCN\textsubscript{10}}, a mid-sized network ($10$ layers, roughly $2.0$M parameters) identical to SMCN, with the exception that all layers between the two max pooling operations are repeated three times. \emph{SMCN\textsubscript{S}}, a simplified architecture where max pooling was replaced with average pooling, and all dropout layers were removed from the network (thus, the activation functions constitute the only non-linearities in this network). \emph{SMCN\textsubscript{BN}}, in which batch normalization is performed before applying the activation functions. We decided not to use dropout in this architecture, as batch normalization in conjunction with dropout can be problematic \citep{li2018understanding}. Note that since batch normalization negates the effect of any preceding scaling, adaptive scaling should not make a difference here. Finally, we also tested the vanilla SMCN with a Stochastic Gradient Descent optimizer with Momentum, instead of the Adam optimizer. Here, the networks were again trained for $\num{60000}$ steps, with the momentum parameter set to $0.9$, an initial learning rate of $\eta=0.01$, and a learning schedule linearly decreasing the learning rate per weight update, reaching $0.0004$ at the end of training.
% parameters
% vanilla SMCN / CIFAR 10: 1'796'672
% vanilla SMCN / CIFAR 100: 1'813'952
% SMCNc / CIFAR10: 2'009'664
% SMCNc / CIFAR100: 2'026'944

\subsection{Performance}

\begin{table}[t]
	\caption{Performance comparison (percentage correct): Common activation functions, adaptive scaling, and ABUs by task, optimizer and network architecture. Table shows mean values of 30 runs plus standard errors per configuration, as well as mean rank across all six configurations. Highest performing activation function per column in bold.}
	\label{performance table 1}
	\begin{center}
    	\resizebox{\textwidth}{!}{% Resize table to fit within \linewidth horizontally
		\begin{tabular}{r|cccccc|c}
			\multicolumn{1}{r|}{\bf Network}
            &\multicolumn{1}{c}{SMCN}
            &\multicolumn{1}{c}{SMCN\textsubscript{10}}
            &\multicolumn{1}{c}{SMCN\textsubscript{S}}
            &\multicolumn{1}{c}{SMCN\textsubscript{BN}}
            &\multicolumn{1}{c}{SMCN}
			&\multicolumn{1}{c}{SMCN}
			&\multicolumn{1}{|c}{}\\
			\multicolumn{1}{r|}{\bf Optimizer}
            &\multicolumn{1}{c}{Adam}
            &\multicolumn{1}{c}{Adam}
            &\multicolumn{1}{c}{Adam}
            &\multicolumn{1}{c}{Adam}
            &\multicolumn{1}{c}{Momentum}
            &\multicolumn{1}{c}{Adam}
            &\multicolumn{1}{|c}{Mean} \\
			\multicolumn{1}{r|}{\bf Task}
            &\multicolumn{1}{c}{CIFAR10}
            &\multicolumn{1}{c}{CIFAR10}
            &\multicolumn{1}{c}{CIFAR10}
            &\multicolumn{1}{c}{CIFAR10}
            &\multicolumn{1}{c}{CIFAR10}
            &\multicolumn{1}{c}{CIFAR100}
            &\multicolumn{1}{|c}{Rank} \\
			\hline
			I   			& $75.51\pm0.11$ & $73.19\pm0.37$ & $38.87\pm0.08$ & $71.72\pm0.09$ & $77.34\pm0.06$ & $44.11\pm0.09$ & $12.00$\\
			$\alpha$I 	 	& $76.52\pm0.08$ & $77.34\pm0.18$ & $39.48\pm0.06$ & $71.34\pm0.10$ & $76.32\pm0.06$ & $45.58\pm0.09$ & $11.50$\\
			\hline
			tanh   			& $75.44\pm0.06$ & $58.55\pm4.47$ & $67.19\pm0.11$ & $75.10\pm0.07$ & $78.76\pm0.05$ & $41.02\pm0.13$ & $12.00$\\
			$\alpha$tanh 	& $79.07\pm0.07$ & $73.40\pm3.87$ & $68.82\pm0.10$ & $75.32\pm0.10$ & $79.14\pm0.05$ & $46.85\pm0.08$ & $9.83$\\
			\hline
			ReLU      		& $79.42\pm0.17$ & $81.07\pm0.15$ & $72.79\pm0.16$ & $81.17\pm0.06$ & $81.63\pm0.07$ & $43.66\pm0.10$ & $8.17$\\
			$\alpha$ReLU    & $79.23\pm0.15$ & $82.97\pm0.12$ & $73.89\pm0.10$ & $81.12\pm0.11$ & $81.85\pm0.07$ & $46.22\pm0.11$ & $7.17$\\
			\hline
			ELU    			& $81.78\pm0.06$ & $83.41\pm0.08$ & $73.33\pm0.13$ & $80.87\pm0.06$ & $82.16\pm0.06$ & $48.59\pm0.11$ & $5.83$\\
			$\alpha$ELU    	& $82.60\pm0.06$ & $84.94\pm0.06$ & $75.03\pm0.13$ & $80.89\pm0.06$ & $82.06\pm0.04$ & $51.03\pm0.10$ & $3.50$\\
			\hline
			SELU    		& $81.75\pm0.07$ & $83.29\pm0.07$ & $71.72\pm0.14$ & $79.36\pm0.05$ & $82.48\pm0.05$ & $48.25\pm0.08$ & $6.83$\\
            $\alpha$SELU    & $82.81\pm0.06$ & $\mathbf{85.04\pm0.04}$ & $73.79\pm0.15$ & $79.57\pm0.07$ & $81.99\pm0.05$ & $51.08\pm0.08$ & $4.33$\\
			\hline
			Swish   		& $82.07\pm0.08$ & $83.73\pm0.07$ & $74.33\pm0.16$ & $\mathbf{81.77\pm0.04}$ & $82.02\pm0.06$ & $49.14\pm0.12$ & $4.33$\\
			$\alpha$Swish   & $82.27\pm0.06$ & $84.56\pm0.05$ & $75.67\pm0.09$ & $81.61\pm0.05$ & $82.35\pm0.05$ & $50.19\pm0.08$ & $3.17$\\
			\hline
			ABU (ours)   	& $\mathbf{83.12\pm0.06}$ & $84.70\pm0.06$ & $\mathbf{76.19\pm0.11}$ & $80.63\pm0.09$ & $\mathbf{83.12\pm0.06}$ & $\mathbf{52.13\pm0.08}$ & $\mathbf{2.33}$\\
		\end{tabular}
        }
	\end{center}
\end{table}

For our performance tests, we chose a vanilla SMCN with Adam optimizer and CIFAR10 as the default setup to compare the various activation functions. All other tested setups are systematically varied versions of this, and differ in only one aspect each, i.e., network architecture, optimizer, or task. On average, adding adaptive scaling yielded improved performance for all activation functions, as evidenced by higher mean ranks of all adaptively scaled activation functions, compared to their fixed counterparts (see Table 1). However, as expected beforehand, batch-normalized networks (SMCN\textsubscript{BN}) were found to be indifferent to adaptive scaling. Interestingly, also in networks trained with the Momentum optimizer, adaptive scaling yielded little to no improvement over the fixed activation functions. Adaptive Blending Units, on the other hand, outperformed all other activation functions in four out of six setups (including the Momentum setup), showing remarkable robustness across architectural choices, and consequently scoring the highest mean rank of all tested activation functions. Since the ability to perform adaptive scaling is an integral part of Adaptive Blending Units, any improvements over adaptively scaled activation functions can likely be attributed to their adaptive shape.

\subsection{Analysis}

\begin{figure}[t]
\centering
\includegraphics[width=1.0\textwidth]{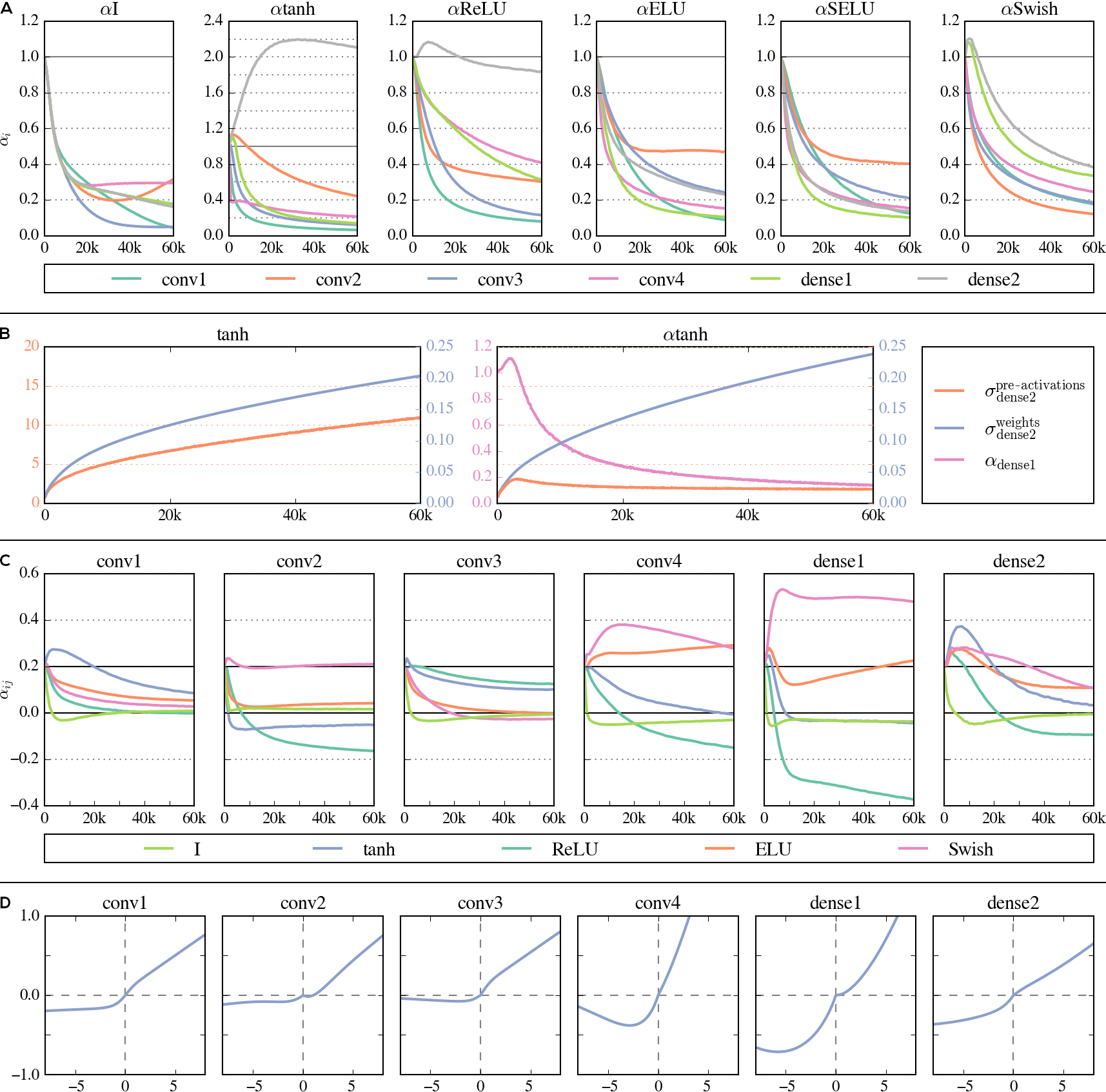}
%\fbox{\rule[-.5cm]{0cm}{4cm} \rule[-.5cm]{4cm}{0cm}}
\caption{\textbf{A}: $\alpha_i$ of adaptively scaled activation functions over the course of training in a vanilla SMCN (mean of $30$ runs, $\num{60000}$ steps). \textbf{B}: Effect of adaptive scaling on pre-activation distributions, exemplified by tanh \& $\alpha$tanh. Scaling weights $\alpha_i$ (magenta) enforce stable variance of next layer's pre-activations (orange), compensating for the increased variance of the regular weight matrix $W_{\text{dense2}}$ (blue). Plots show mean of five runs. \textbf{C}: ABU blending weights $\alpha_{ij}$ over the course of training. \textbf{D}: Average activation functions (ABU) by layer at the end of training (axes scaled to improve readability).}
\end{figure}

But what exactly changes in the networks, when we introduce adaptive scaling or ABUs? In order to provide some insight into the mechanisms introduced by the two approaches, we carried out further analyses based on the default setup, i.e., a vanilla SMCN with Adam optimizer trained on CIFAR10.

Let us first examine how scaling weights behave during training. In our tests, the scaling weights $\alpha_i$ almost unanimously decreased to values far below $1$ (see Figure 2A). This behavior was highly consistent over repeated runs with random initializations and mini-batch sampling: The mean standard deviation (over repeated runs) of the final scaling weights reached at the end of training is $\text{mean}(\sigma_{\alpha_i})=0.023$. With respect to how this influences the activations in the network, it is sensible to consider the succeeding layer's pre-activation statistics, i.e., the distribution of values going into its activation function: The distribution of pre-activation is approximately Gaussian for large layers due to the Central Limit Theorem, and is thus easier to compare between networks with different activation functions. For many activation functions, the pre-activations are also crucial with respect to the magnitude of the gradients, in that they determine the fraction of inputs reaching saturated regions of the activation function. Our analyses show that the decreasing scaling weights rather precisely counteract an increase in the variance of the following weight matrix over the course of training. This stabilizes the distributions of pre-activation states in the following layers in both mean and variance, thus drastically reducing any covariate shift. We illustrate this by comparing the pre-activation variance of the last layer in SMCN networks, using tanh and $\alpha$tanh, in Figure 2B. Without adaptive scaling, the variance of pre-activations increased throughout training for all layers and all tested activation functions. With adaptive scaling, the standard deviations typically converged to a value between $0.5$ and $5$ early on in training, and remained stable at this value for the remainder of the training procedure. At the same time, pre-activation means were kept stable at less than a standard deviation from zero. 

We take from this that adaptive scaling acts as a normalization technique, similar to batch normalization \citep{ioffe2015batch} or layer normalization \citep{ba2016layer}. In contrast to these, however, adaptive scaling doesn't require any explicit calculations of variance or other statistics, or keeping track of running averages in inference, and does not depend on batch or layer size. In principle, it also allows the network to optimize the statistics of the neurons' input distributions, instead of enforcing unit mean and variance across the layer or batch. Depending on the type of calculations that are predominantly performed by a given layer, it appears plausible that some layers would prefer stronger saturation, while others may benefit from less saturation, provided the activation functions feature increasingly saturating regions. That being said, our analysis does not allow us to infer whether or not the realized distributions actually constitute an optimum for the required computation in a given layer. If an activation function is a homogeneous function of degree $1$ (scale-invariant; e.g., ReLU), the network performance would likely not be influenced by the variance of the pre-activation's distribution, but may still be affected by shifts of the mean, which are also mitigated by adaptive scaling. We consider an in-depth analysis of such self-organizing processes, as well as further exploration of this principle for deep networks highly desirable, but out of scope for this work.

Turning to ABUs, we observe the same normalizing effect on the pre-activation statistics of succeeding layers. As illustrated in Figure 2C, ABUs realize a layer's overall downscaling in multiple ways. In the first convolutional layer, for instance, the weights unanimously decrease and mostly converge towards values close to zero. At the end of training, the identity and ReLU have arrived at effectively zero, while the final activation function is mostly a mixture of ELU and tanh. By contrast, the first dense layer achieves the overall downscaling of positive inputs by subtracting ReLU from a mixture of ELU and Swish. In both cases, the resulting function is rather flat, pushing the activations (layer output) closer to zero. These different compositions of blending weights translate into substantially different shapes of the resulting ABU (see Figure 2D). But while the variation of the ABUs' shape across layers is substantial, their shape within each layer is remarkably consistent over repeated runs, as indicated by a mean standard deviation of $\sigma^{\alpha_{ij}}_{\text{mean}}=0.010$ per layer and blending weight. This consistency, in conjunction with the good performance figures achieved by ABUs, lead to the conclusion that the learned shapes are meaningful with respect to the computations performed in the network.

In summary, while adaptive scaling stabilizes the pre-activation statistics of succeeding layers, the learned shapes of the resulting functions are meaningful, as well. Moreover, both adaptive scaling and an adaptive shape were found to yield improvements in performance for image classification tasks with convolutional networks.

\section{Normalized Blending Weights}

\begin{table}[t]
	\caption{Performance comparison (percentage correct): ABU and the normalized ABU\textsubscript{abs}, ABU\textsubscript{nrm} ABU\textsubscript{pos}, and ABU\textsubscript{soft} by task, optimizer and network architecture. Table shows mean values of 30 runs plus standard errors per configuration, as well as mean rank across all six configurations. Highest performing activation function per column in bold.}
	\label{performance table 2}
	\begin{center}
    	\resizebox{\textwidth}{!}{% Resize table to fit within \linewidth horizontally
		\begin{tabular}{r|cccccc|c}
			\multicolumn{1}{r|}{\bf Network}
            &\multicolumn{1}{c}{SMCN}
            &\multicolumn{1}{c}{SMCN\textsubscript{10}}
            &\multicolumn{1}{c}{SMCN\textsubscript{S}}
            &\multicolumn{1}{c}{SMCN\textsubscript{BN}}
            &\multicolumn{1}{c}{SMCN}
			&\multicolumn{1}{c}{SMCN}
			&\multicolumn{1}{|c}{}\\
			\multicolumn{1}{r|}{\bf Optimizer}
            &\multicolumn{1}{c}{Adam}
            &\multicolumn{1}{c}{Adam}
            &\multicolumn{1}{c}{Adam}
            &\multicolumn{1}{c}{Adam}
            &\multicolumn{1}{c}{Momentum}
            &\multicolumn{1}{c}{Adam}
            &\multicolumn{1}{|c}{Mean} \\
			\multicolumn{1}{r|}{\bf Task}
            &\multicolumn{1}{c}{CIFAR10}
            &\multicolumn{1}{c}{CIFAR10}
            &\multicolumn{1}{c}{CIFAR10}
            &\multicolumn{1}{c}{CIFAR10}
            &\multicolumn{1}{c}{CIFAR10}
            &\multicolumn{1}{c}{CIFAR100}
            &\multicolumn{1}{|c}{Rank} \\
			\hline
			ABU   					& $83.12\pm0.06$ & $84.70\pm0.06$ & $\mathbf{76.19\pm0.11}$ & $80.63\pm0.09$ & $83.12\pm0.06$ & $\mathbf{52.17\pm0.08}$ & $\mathbf{2.5}$\\
			ABU\textsubscript{nrm}	& $82.82\pm0.07$ & $84.18\pm0.07$ & $75.99\pm0.10$ & $81.39\pm0.07$ & $\mathbf{83.29\pm0.06}$ & $51.88\pm0.10$ & $3.7$\\
			ABU\textsubscript{abs}	& $\mathbf{83.14\pm0.08}$ & $\mathbf{84.95\pm0.06}$ & $76.07\pm0.16$ & $81.17\pm0.08$ & $82.32\pm0.05$ & $52.16\pm0.07$ & $2.7$\\
            ABU\textsubscript{pos}	& $82.90\pm0.05$ & $84.05\pm0.06$ & $76.10\pm0.11$ & $81.44\pm0.07$ & $83.26\pm0.06$ & $51.90\pm0.09$ & $3.2$\\
            ABU\textsubscript{soft}	& $82.54\pm0.07$ & $84.63\pm0.05$ & $76.18\pm0.07$ & $\mathbf{81.51\pm0.06}$ & $82.09\pm0.05$ & $52.10\pm0.08$ & $3.0$\\
		\end{tabular}
        }
	\end{center}
\end{table}

So far, we focused on adaptive scaling as an integral part of ABUs. In order to better understand the effects of shape in ABUs, we conducted an additional experiment, in which we normalized the blending weights of the ABUs in four different ways, taking away their ability to scale the layer output by overall increases or decreases of the blending weights.

\subsection{Methods}

The following four methods of normalization for ABUs were used: \emph{ABU\textsubscript{nrm}} denotes the case where a layer's blending weights are normalized to sum up to $1$ ($\sum_j \alpha_{ij} = 1$). The normalization was implemented as part of the graph, dividing the blending weights by their sum, before applying them in the respective ABU. Similarly, in \emph{ABU\textsubscript{abs}}, we divided the raw blending weights by the sum of their absolute values, thus keeping the sum of the absolute values of the blending weights at $1$ ($\sum_j |\alpha_{ij}| = 1$). Note that under this constraint, scaling is still possible, albeit not independent of the resulting shape: By having similar activation functions cancel each other out with blending weights on either side of zero, functions can be constructed that return only a fraction of the input, or even zero, for all positive values. We decided to include this form of normalization in the test to provide a more complete account of possible normalizations. In \emph{ABU\textsubscript{pos}}, any negative values are clipped before normalization, such that all blending weights are strictly positive ($\sum_j \alpha_{ij} = 1$; $\alpha_{ij} > 0$). Finally, in \emph{ABU\textsubscript{soft}}, we realized the same constraint (all-positive and normalized) by applying softmax normalization to the blending weights. With the exception of ABU\textsubscript{abs}, none of the normalized versions of ABUs can realize an overall scaling of the resulting functions for positive input values. For the experiments, we used the same network and task configurations as in the previous section.

\subsection{Performance \& Analysis}

The results of our performance tests are reported in Table 2. All five versions of ABUs showed remarkably similar performance throughout the tested task and network configurations - the average gap between the best and weakest performing ABU in a given setup is a mere $0.63$\%. In terms of mean rank, ABU and ABU\textsubscript{abs} lead the field, and are thus the two most robust configurations. However, none of the other three versions fell far behind. We again used the default setup (vanilla SMCN, Adam, CIFAR10) for an analysis of the blending weights and their effects on the succeeding pre-activations. We found ABU\textsubscript{abs}s to behave much like unconstrained ABUs, implementing adaptive scaling, keeping the layer statistics constant over the course of training, thus mitigating covariate shift. Despite the fact that the scaling imposes constraints on the shape of the resulting function (as outlined above), ABU\textsubscript{abs} performed very similarly to unconstrained ABUs in most settings. By contrast, but very much expectedly, ABU\textsubscript{nrm}, ABU\textsubscript{pos}, and ABU\textsubscript{soft}, were unable to keep the layer statistics at constant levels, and a considerable covariate shift akin to that in fixed activation functions was observed. Interestingly, this appears to have only a minor impact on performance, and they were able to keep up with, or even outperform unconstrained ABUs in some of the tested settings. The good performance of normalized ABUs in our tests is in line with \cite{manessi2018learning}, who found very similar or identical units\footnote{With respect to the constraints, \citet{manessi2018learning}'s \emph{affine()} units are equivalent to ABU\textsubscript{nrm}, and their \emph{convex()} units are equivalent to ABU\textsubscript{pos}. Unfortunately, the authors did not provide details with respect to their implementation, so we cannot say whether or not the implementations are equivalent.} to outperform common activation functions in MNIST, CIFAR and ImageNet tasks, using widely used network architectures, such as AlexNet and ResNet-56.
In conclusion, while ABUs generally apply adaptive scaling when possible, the ability to learn the function's shape by itself already helps to improve network performance beyond the level of the established activation functions they are comprised of.

\section{Pre-Training Scaling and Blending weights}

\begin{table}[t]
	\caption{Performance (percentage correct) after initialization of scaling / blending weights $\alpha$ at the final values of a preceding run, i.e., after $\num{60000}$ steps. Additionally for ABUs: Pre-trained weights normalized at initialization to avoid too low initial scaling. Highest performing treatment of scaling and blending weights per activation function (row) indicated in bold.}
	\label{pre-trained weights performance}
	\begin{center}
    	\resizebox{\textwidth}{!}{% Resize table to fit within \linewidth horizontally
		\begin{tabular}{r|c|cccc}
			\multicolumn{1}{r|}{\bf Network}
            &\multicolumn{1}{c|}{SMCN}
			&\multicolumn{1}{c}{SMCN}
            &\multicolumn{1}{c}{SMCN}
            &\multicolumn{1}{c}{SMCN}
            &\multicolumn{1}{c}{SMCN}\\
			\multicolumn{1}{r|}{\bf Optimizer}
            &\multicolumn{1}{c|}{Adam}
            &\multicolumn{1}{c}{Adam}
            &\multicolumn{1}{c}{Adam}
            &\multicolumn{1}{c}{Adam}
            &\multicolumn{1}{c}{Adam}\\
			\multicolumn{1}{r|}{\bf Task}
            &\multicolumn{1}{c|}{CIFAR10}
            &\multicolumn{1}{c}{CIFAR10}
            &\multicolumn{1}{c}{CIFAR10}
            &\multicolumn{1}{c}{CIFAR10}
            &\multicolumn{1}{c}{CIFAR10}\\
			\multicolumn{1}{r|}{\bf $\alpha$ init}
            &\multicolumn{1}{c|}{$\frac{1}{m}$}
            &\multicolumn{1}{c}{pre-trained}
            &\multicolumn{1}{c}{pre-trained}
            &\multicolumn{1}{c}{pre-trained (norm.)}
            &\multicolumn{1}{c}{pre-trained (norm.)} \\
			\multicolumn{1}{r|}{\bf $\alpha$ trainable}
            &\multicolumn{1}{c|}{\checkmark}
            &\multicolumn{1}{c}-
            &\multicolumn{1}{c}{\checkmark}
            &\multicolumn{1}{c}-
            &\multicolumn{1}{c}{\checkmark}\\
			\hline
			$\alpha$tanh 	 		& $79.07\pm0.07$ & $79.60\pm0.07$ & $\mathbf{79.71\pm0.06}$ & - & - \\
			$\alpha$ReLU    		& $79.23\pm0.15$ & $79.62\pm0.11$ & $\mathbf{79.77\pm0.09}$ & - & - \\
			$\alpha$ELU    			& $\mathbf{82.60\pm0.06}$ & $79.45\pm0.15$ & $81.22\pm0.10$ & - & - \\
			\hline
			ABU   					& $\mathbf{83.12\pm0.06}$ & $80.93\pm0.15$ & $82.02\pm0.15$ & $80.99\pm0.12$ & $82.88\pm0.06$ \\
			ABU\textsubscript{nrm}	& $\mathbf{82.82\pm0.07}$ & $78.88\pm0.08$ & $81.78\pm0.14$ & - & - \\
			ABU\textsubscript{abs}	& $\mathbf{83.14\pm0.08}$ & $79.80\pm0.24$ & $82.42\pm0.21$ & - & - \\
			ABU\textsubscript{pos}	& $\mathbf{82.90\pm0.05}$ & $78.93\pm0.07$ & $81.45\pm0.16$ & - & - \\
			ABU\textsubscript{soft}	& $82.54\pm0.07$ & $\mathbf{82.69\pm0.05}$ & $81.69\pm0.08$ & - & - \\
		\end{tabular}
        }
	\end{center}
\end{table}

Finally, we investigated for both adaptive scaling and ABUs, whether or not the adaptiveness of scaling and blending weights by itself is an important factor for the overall performance of the network, and if the performance could possibly be further improved by using pre-trained weights. To this end, we set up an experiment with two main conditions. In both of them, we initialized the networks with the final scaling or blending weights of a preceding run. We then fixed these values after initialization in one condition, while keeping them adaptive in another. In case of ABUs, it is conceivable that the \emph{shape} of the function at the end of training would be ideal, while the \emph{scaling} may be too low at the beginning of a new run. Therefore, we added two more conditions akin to the main condition and only for unconstrained ABUs, in which we normalized the pre-trained blending weights after initialization, thus keeping the learned shape, while resetting the learned scaling. All tests were based on the default setup (vanilla SMCN, Adam, CIFAR10).

The results are shown in Table 3. For $\alpha$tanh and $\alpha$ReLU, initializing the scaling weights at the predominantly low final values of a preceding run helped to improve performance. In both cases, runs with fixed pre-trained $\alpha_i$ already surpassed the performance of the preceding run, but keeping them adaptive over the course of training led to further improvements. The fact that fixed pre-trained scaling weights yielded an increase in performance suggests that the initial variance of weights in the weight matrices (derived using He initialization), may not have been ideal as initial conditions, despite resulting in pre-activation variances of about $1$. $\alpha$ELU, by contrast, substantially lost performance after initialization with pre-trained scaling weights, irrespective of whether or not they were fixed or adaptive throughout the run. With the exception of ABU\textsubscript{soft}, the same was the case in all versions of ABUs, where fixed blending weights, in particular, led to sizable drops in performance of up to four percent. ABU\textsubscript{soft}, being the exception to this rule, improved slightly for fixed pre-trained blending weights, but lost performance with adaptive pre-trained blending weights. Normalizing the unrestricted ABU blending weights after initialization with pre-trained values led to improvements over non-normalized pre-trained blending weights, but the default setup with initialization at $\frac{1}{m}$ still performed best. Overall, we found all but one of the tested activation functions to perform best, when the blending weights were kept adaptive, as opposed to fixed after initialization. These results suggest that in both adaptive scaling and ABUs, much of the gained performance is won by keeping the scaling and / or shape adaptive. Moreover, the fact that this applies also to most normalized versions of ABUs indicates that there may not be any single optimal shape for an activation function in a given layer.

\section{Conclusion \& Outlook}

In summary, we introduced Adaptive Blending Units (ABUs), and analyzed adaptive scaling for common activation functions. We found robust performance advantages of both approaches over established activation functions in a range of tasks and network architectures. In adaptive scaling, the performance advantages could be traced back to stabilized pre-activation statistics during training, mitigating covariate shift. The same behavior was found for unconstrained ABUs, while normalized ABUs reached similar levels of performance without the ability to significantly scale the layer output. Our results suggest that the adaptiveness of the shape over the course of training may play a major role in this, as well, as opposed to simply converging to some ideal shape.

With respect to adaptive scaling, a logical next step would be to introduce a shifting parameter per layer, to allow the network to further optimize the input distributions to the activation functions, and to move this learned normalization in front of the activation: $f(\alpha_i\cdot(x+\beta_i))$. This form of self-organized normalization could be explicitly combined with ABUs, thus detaching the handling of layer statistics from the shape of the activation function. Recurrent networks may be a particularly interesting field of application for self-optimization of layer statistics, as it should, in principle, mitigate some of the issues associated with explicit normalization techniques. Interestingly, adaptive scaling has been discussed for neural populations outside of the field of deep learning and proven helpful in maintaining stable output distributions \citep{turrigiano2004homeostatic, leugering2018unifying}. Beyond this, an increase in the number of distinct ABUs within a layer may yield further improvements in performance, as well as a systematic search for high-performing sets of activation functions in ABUs.

\section*{Acknowledgments}
We would like to thank the Nvidia Corporation for their kind donation of a Titan X Pascal graphics card, as well as Tom Hatton for his assistance in this project.

\bibliography{abu_citations}
\bibliographystyle{iclr2017_conference}

\end{document}